\documentclass[a4paper,twoside]{article}

\pdfoutput=1

\usepackage{epsfig}
\usepackage{subfigure}
\usepackage{calc}
\usepackage{amssymb}
\usepackage{amstext}
\usepackage{amsmath}
\usepackage{amsthm}
\usepackage{multicol}
\usepackage{pslatex}
\usepackage{apalike}

\usepackage{booktabs}
\usepackage{cite}
\usepackage{tabularx}
\usepackage{SCITEPRESS}     

\newcommand{\zalando}[0]{Zalando}
\newcommand{\seb}[0]{Sebastian Heinz}
\newcommand{\christian}[0]{Christian Bracher}
\newcommand{\FashionDNA}[0]{FashionDNA}
\newcommand{\fDNA}[0]{fDNA}
\newcommand{\Europe}[0]{Europe}
\newcommand{\longname}{Studio2Shop}

\def\ie{\emph{i.e.}\ }
\def\etal{\emph{et al}\ }

\makeatletter
\let\@fnsymbol\@arabic
\makeatother

\begin{document}

\title{\longname: from studio photo shoots to fashion articles}

\author{\authorname{Julia Lasserre\thanks{These authors contributed equally.}, Katharina Rasch\footnotemark[1] and Roland Vollgraf}
\affiliation{Zalando Research, Muehlenstr. 25, 10243 Berlin, Germany}
\email{julia.lasserre@zalando.de}
}


\abstract{Fashion is an increasingly important topic in computer vision, in particular the so-called \emph{street-to-shop} task of matching street images with shop images containing similar fashion items. Solving this problem promises new means of making fashion searchable and helping shoppers find the articles they are looking for. This paper focuses on finding pieces of clothing worn by a person in full-body or half-body images with neutral backgrounds. Such images are ubiquitous on the web and in fashion blogs, and are typically studio photos, we refer to this setting as \emph{studio-to-shop}. Recent advances in computational fashion include the development of domain-specific numerical representations. Our model Studio2Shop builds on top of such representations and uses a deep convolutional network trained to match a query image to the numerical feature vectors of all the articles annotated in this image. Top-$k$ retrieval evaluation on test query images shows that the correct items are most often found within a range that is sufficiently small for building realistic visual search engines for the studio-to-shop setting.}

\onecolumn \maketitle \normalsize \vfill

\section{\uppercase{INTRODUCTION}}
\label{sec:intro}

Online fashion is a fast growing field which generates massive amounts of (largely unexplored) data. The last five years have seen an increasing number of academic studies specific to fashion in top conferences, and particularly in the computer vision community, with topics as varied as article tagging \cite{Chen2012, Bossard2012, Chen2015}, clothing parsing \cite{Wang2011, Yamaguchi2012, Dong2013, Yamaguchi2013, Liu2014}, article recognition \cite{Wang2011b, Fu2012, Liu2012, Kalantidis2013, Huang2015, Liu2016}, style recommenders (magic mirrors or closets) \cite{Liu2012b, Di2013, Kiapour2014, Jagadeesh2014, Vittayakorn2015, Yamaguchi2015}, and fashion-specific feature representations \cite{Bracher2016, Simoserra2016}. Our company \zalando\ is \Europe's leading online fashion platform and, like other big players on the market, benefits from large datasets and has a strong interest in taking part in this effort.\footnote[2]{This paper is best viewed in colour.}
\begin{figure}[htb!]
  \centering
  \subfigure[Full-body with occlusion]{\includegraphics[width=0.13\textwidth]{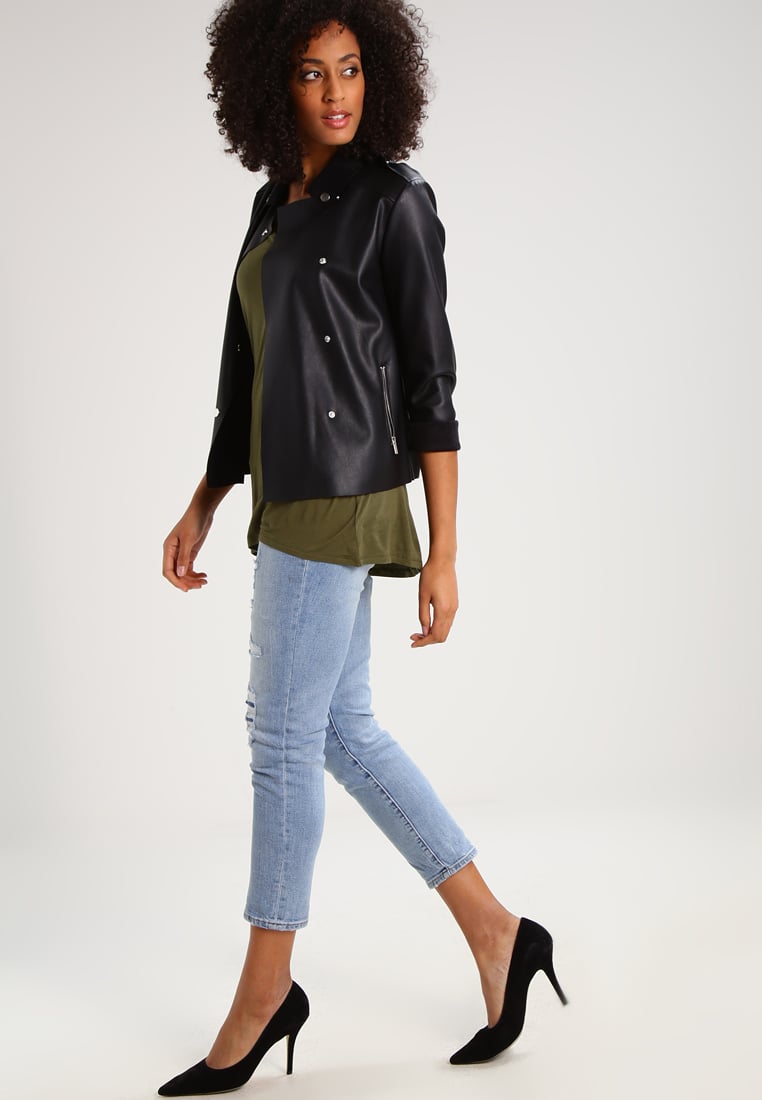}}
  \hfill
  \subfigure[Full-body]{\includegraphics[width=0.13\textwidth]{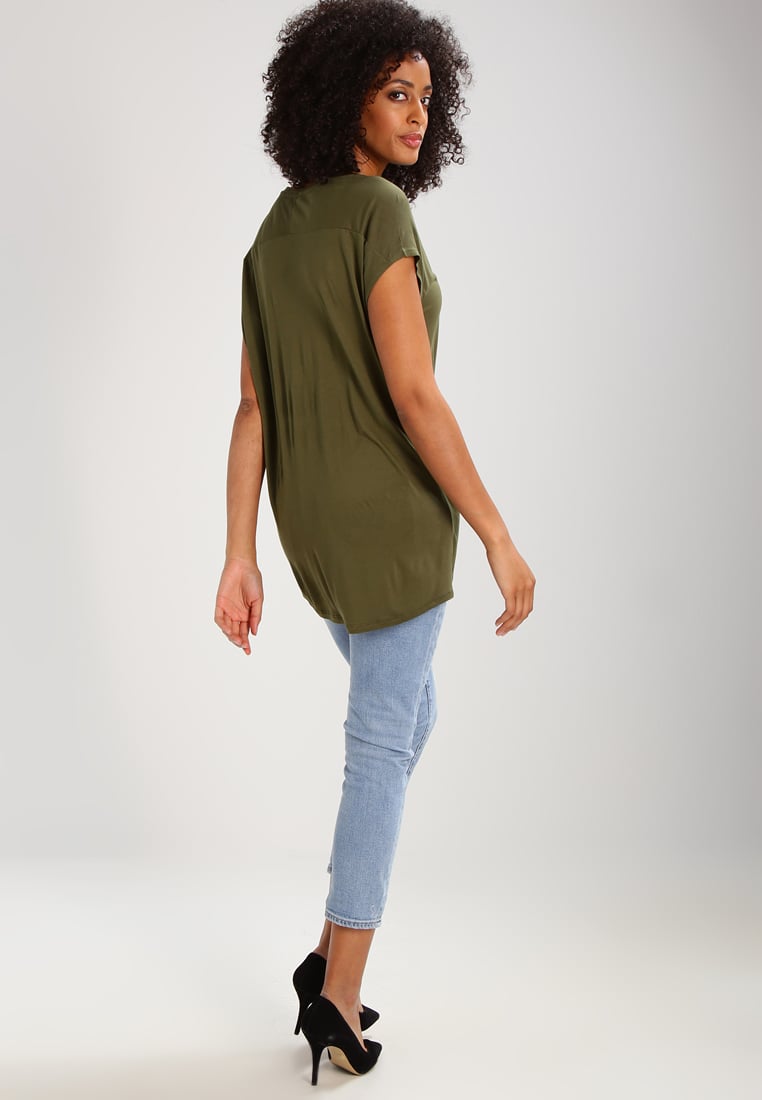}}
  \hfill
  \subfigure[Half-body]{\includegraphics[width=0.13\textwidth]{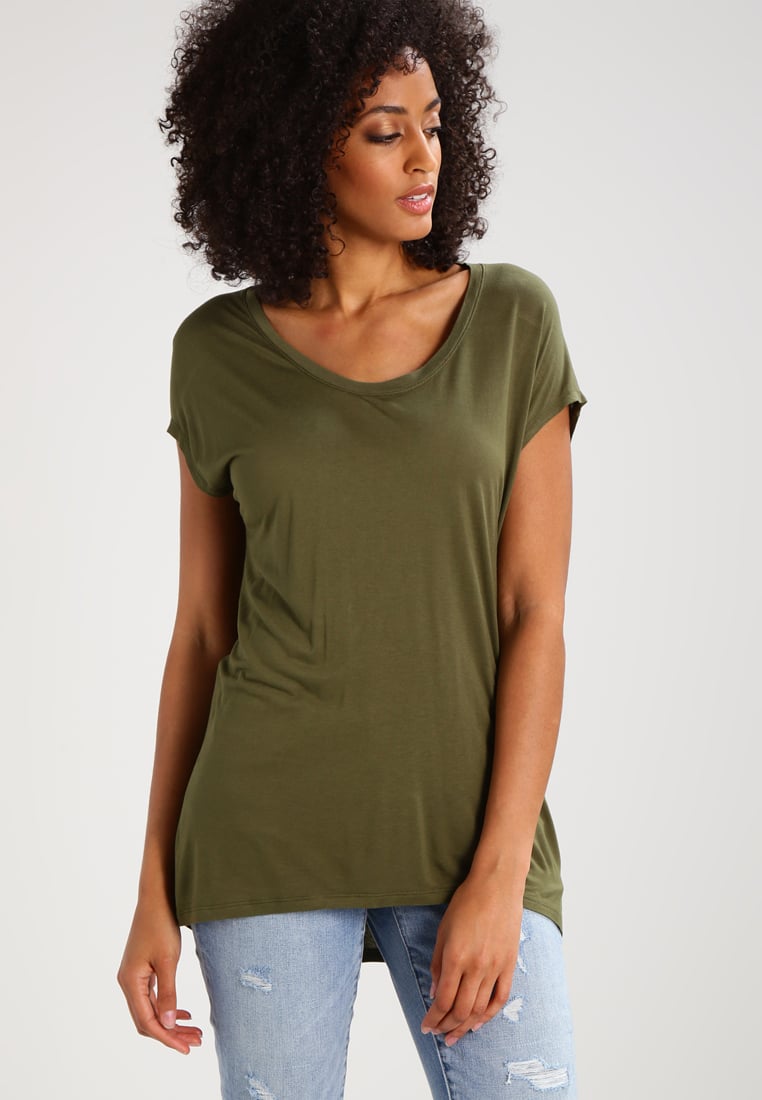}}
  \hfill\\
  \hspace{1cm}
  \subfigure[Detail]{\includegraphics[width=0.13\textwidth]{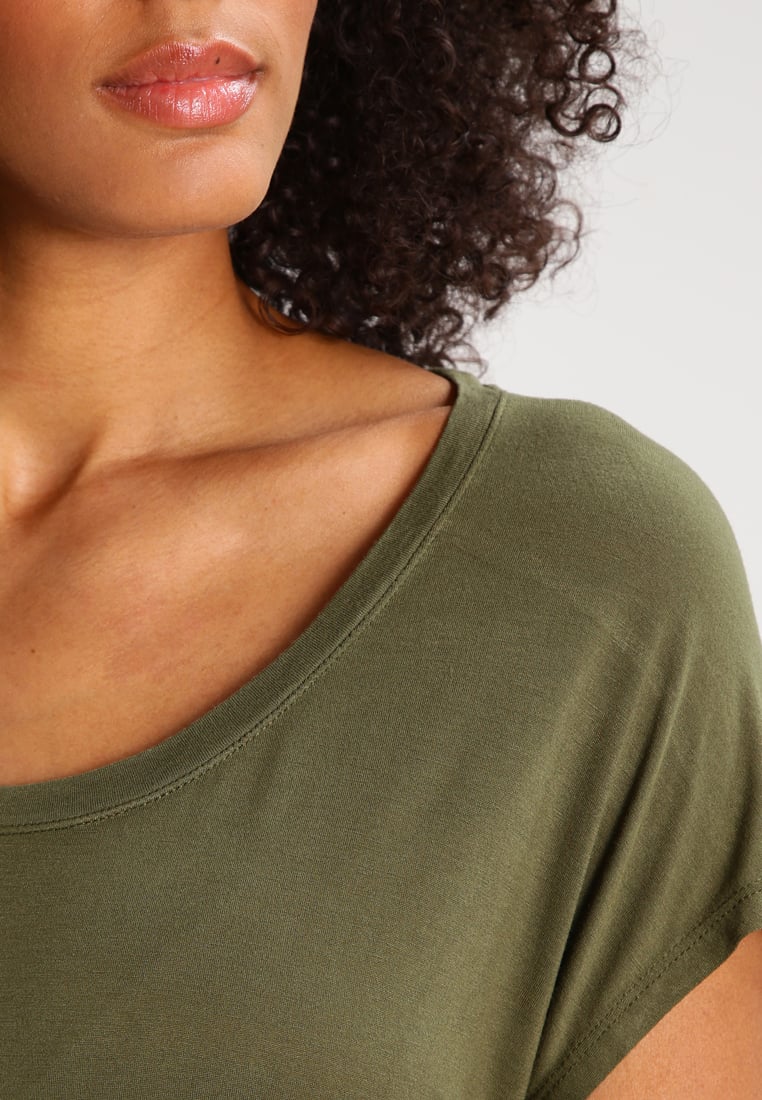}}
  \hfill
  \subfigure[Title]{\includegraphics[width=0.13\textwidth]{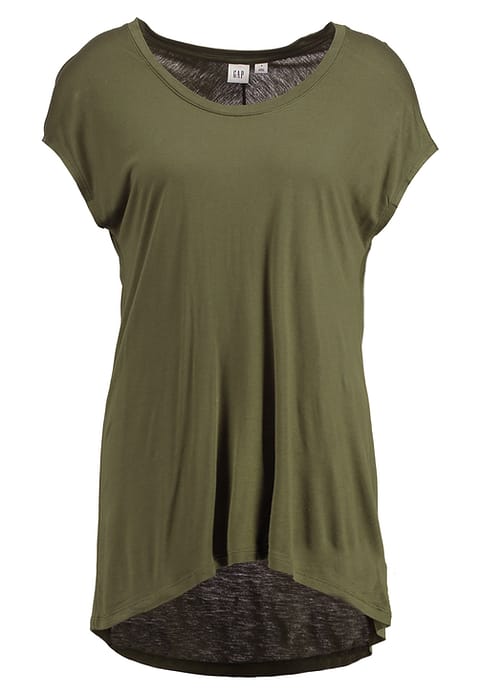}}
  \hfill
  \caption{Examples of images in our dataset. Image types (a-d) are query images featuring models, image type (e) represents the articles we retrieve from.}
  \label{fig:images}
\end{figure}
\begin{table*}[htb!]
  \begin{center}
    \caption{Overview of street-to-shop studies. CV stands for classical computer vision, CNN for convolutional neural network.}
    \resizebox{\textwidth}{!}{%
    \begin{tabular}{ll|lll|lll|l} \toprule
						&			& \multicolumn{3}{c|}{domain 1}						& \multicolumn{3}{c|}{domain 2} & \\
						& dataset 		& focus			& background	& assumptions			& focus				& background	& assumptions & method \\ \midrule
      Street-To-Shop \cite{Liu2012}		& Street-To-Shop	& half + full		& yes		& body-part detector		& half + full			& no	& body-part detector & CV \\
						&			&			&		& upper vs lower		&				&	& upper vs lower & \\ \midrule
      Where-To-Buy-It \cite{Kiapour2015}	& Exact-Street-To-Shop	& half + full		& yes		& category of item		& half + full + title		& no	& 	& CV \\
						&			&			&		& bounding box			&				&	&	& \\ \midrule
      DARN \cite{Huang2015}			& DARN dataset		& half, frontal view	& yes		& upper clothing		& half + title, frontal view	& yes+no& upper clothing & siamese CNN \\
						&			&			&		& frontal view			&				&	& frontal view & \\
						&			&			&		& clothing detector to crop	&				&	& clothing detector to crop & \\ \midrule
      Wang \etal \cite{Wang2016}		& Exact-Street-To-Shop	& half + full		& yes		& none				& half + full + title		& no	& none	& siamese CNN \\
						& AliBaba		&			&		&				&				&	&	& shared lower layers \\ \midrule
      DeepFashion \cite{Liu2016a}		& In-Shop		& half + full + title	& no		& landmark (training)		& \multicolumn{3}{c|}{\textit{same as domain 1}} & single CNN \\ \midrule
      DeepFashion \cite{Liu2016a}		& Consumer-To-Shop	& half + full + title	& yes		& landmark (training)		& \multicolumn{3}{c|}{\textit{same as domain 1}} & single CNN \\  \midrule
      \longname					& \zalando		& half + full		& no		& none				& title				& no	& none & siamese CNN \\
						&			&			&		&				&				&	&	& second input as features \\
      \bottomrule
    \end{tabular}}
    \label{tab:state-of-art}
  \end{center}
\end{table*}

\emph{Street-to-shop} \cite{Liu2012} is the task of retrieving articles from a given assortment that are similar to articles in a query picture from an unknown source (a photo taken on the street, a selfie, a professional photo). In this study, we follow the \emph{exact-street-to-shop} variant \cite{Kiapour2015}, where query images are related to the assortment and therefore the exact-matching article should be retrieved. In particular, we focus on a setting we call \emph{studio-to-shop} where the query picture is a photo shoot image of a model wearing one or several fashion items in a controlled setting with a neutral background, and where the target is an article of our assortment. Such images are ubiquitous on the web, for example in fashion magazines or on online shopping websites, and solving our task would allow our customers to search for products more easily. In addition, this setting can have many internal applications such as helping trend scouts match blog images with our products, or annotating all our catalogue images with all contained articles automatically.

Recognising fashion articles is a challenging task. Clothing items are not rigid and usually undergo strong deformations in images, they may also be partially occluded. They vary in most of their physical attributes (for example colour, texture, pattern), even within the same category, and can contain details of importance such as a small logo, so that recovering a few basic attributes may not always be enough. Our query images contain a wide variety of positions and views, with full-body and half-body model images, as well as images of details (see Figures~\ref{fig:images}(a-d)).

In the literature, street-to-shop is typically approached as an isolated problem, taking pairs of query/target images together with optional article meta-data as input. So far, both query and target inputs have had the same form, namely a feature vector or an image. By putting forward our model \longname, we propose instead to build on existing feature representations of fashion articles, and to use images for queries and static feature vectors for targets. Indeed, many online shops, including \zalando, already have a well-tuned feature representation of their articles that can be used across a wide array of applications, including recommender systems and visual search. Moreover, such features are also publicly available (e.g.\ AlexNet's fc6 \cite{Krizhevsky2012}, VGG16's fc14 \cite{Simonyan2014}, fashion in 128 floats \cite{Simoserra2016}).

Breaking the symmetry between query and target allows us to use more complete feature representations, since static features have usually been trained on massive datasets. The representation we use in this study was trained on hundreds of thousands of articles, including categories of articles that are not directly relevant to our task, and on more attributes than we could process with an end-to-end framework. Note that only title images were used during training, i.e.\ images of the article without any background as shown in Figure~\ref{fig:images}(e), and no model images were seen.\\

The contribution of our work is three-fold:

\begin{itemize}
  \item We naturally handle all categories at hand and make no assumptions about the format of the model image (full body, half body, detail), nor do we require additional information such as the category, a bounding box or landmarks.
  \item We show that in our setting, end-to-end approaches are not necessary, and that using a static feature representation for the target side is effective, especially when combining it with a non-linear query-article matching model.
  \item We show that we can achieve reasonable results on an extra (publicly available) dataset with similar properties, without even fine-tuning our model.
\end{itemize}

The remainder of this paper is organised as follows: we review the related work in Section~\ref{sec:related}, describe our dataset and approach in detail in Sections~\ref{sec:data} and~\ref{sec:method}, and evaluate our approach in Section~\ref{sec:evaluation}.
\begin{figure*}[htb!]
  \centering
  \hfill
  \subfigure[siamese]{\includegraphics[width=0.25\textwidth]{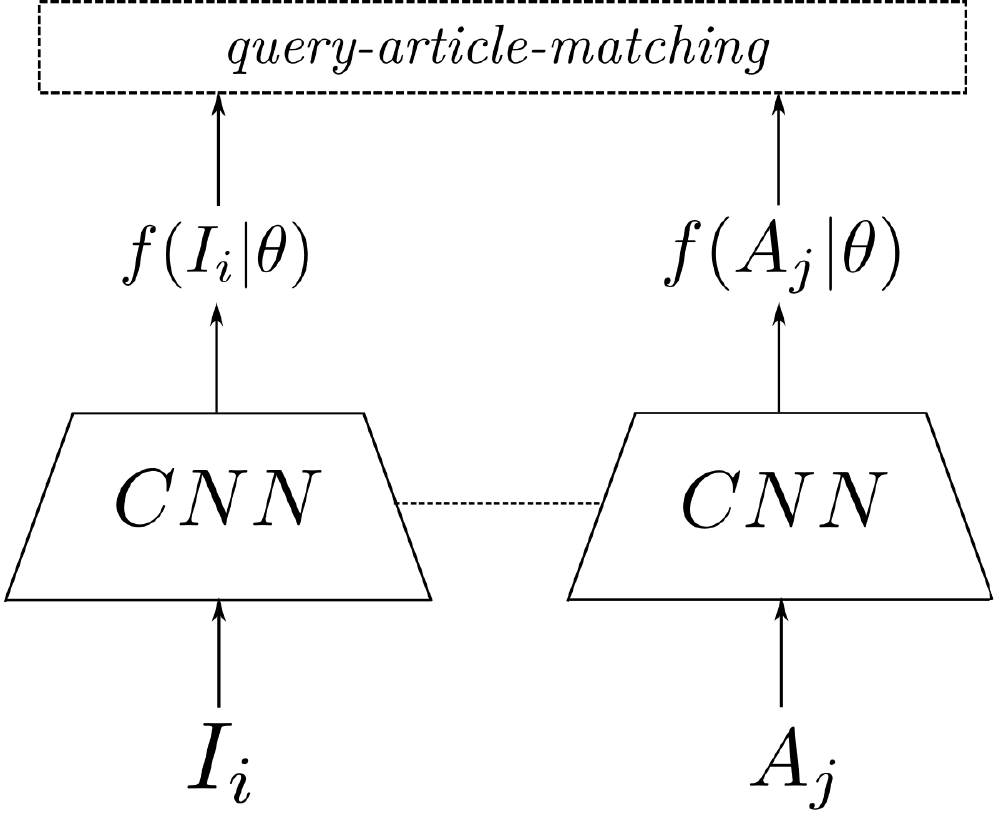}} \hfill
  \subfigure[static]{\includegraphics[width=0.25\textwidth]{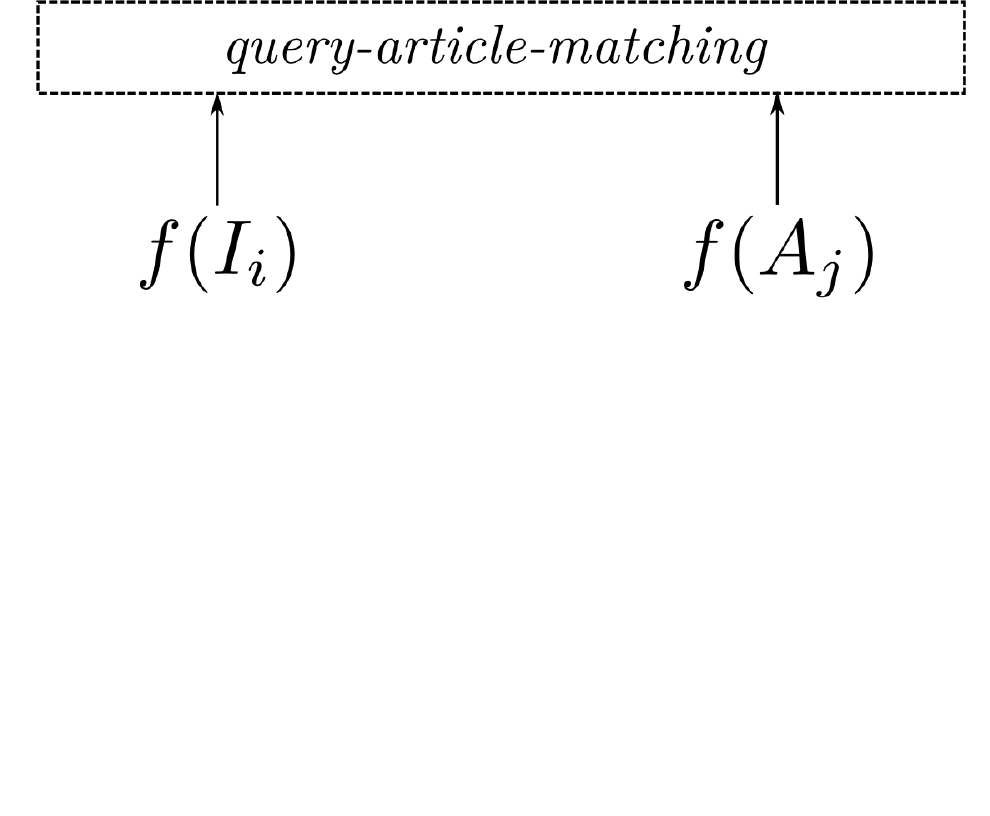}} \hfill
  \subfigure[ours]{\includegraphics[width=0.25\textwidth]{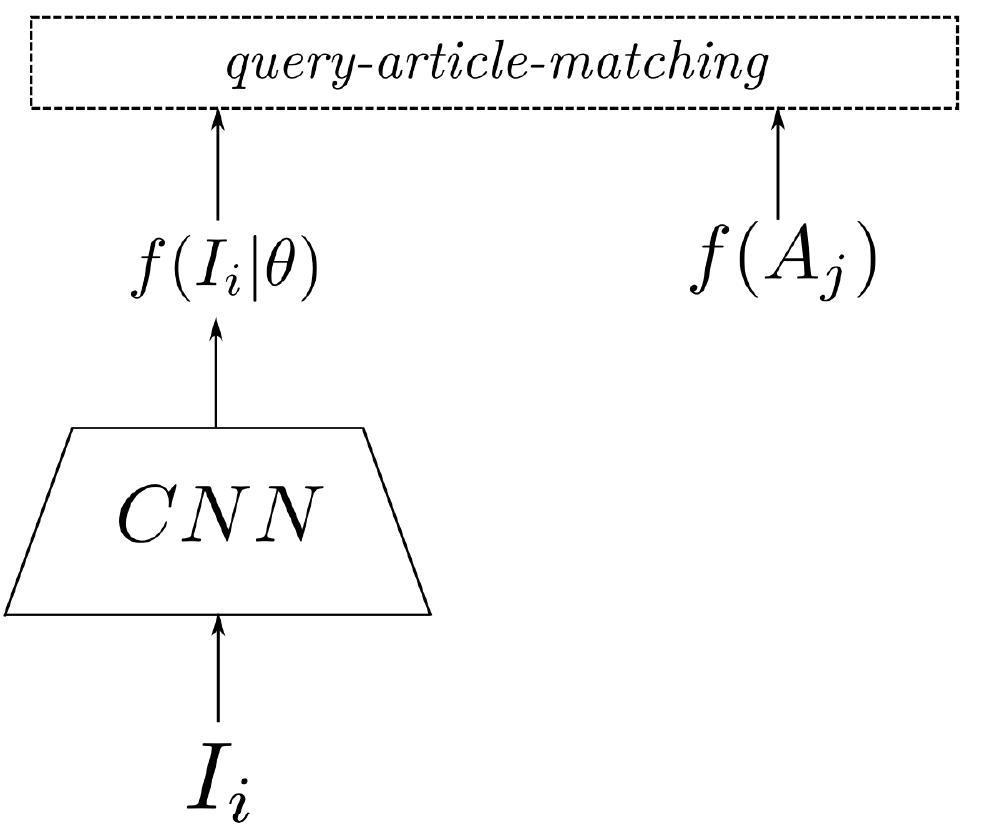}}
  \hfill
  \caption{Variations of street-to-shop architectures in the literature. $\theta$ represents the joint set of parameters of the two legs. (a) The most common architecture in recent studies, based on image pairs. (b) Only static features, as found in \cite{Kiapour2015}. (c) Our approach.}
  \label{fig:lower}
  \hspace{0.5cm}
\vspace{-0.5cm}
\end{figure*}

\section{RELATED WORK}
\label{sec:related} 

\paragraph{Data.} The street-to-shop task was defined in \cite{Liu2012} and formulated as a domain transfer problem. Since then, many groups have contributed their own dataset with their own set of assumptions. There is a lot of variation in the literature regarding types of street (query) and shop (target) images. In order to reduce background, street images are often assumed to be cropped around the article \cite{Kiapour2015, Huang2015} and sometimes come with the information of the category of the article \cite{Kiapour2015}. Shop images vary across and within datasets, often mixing full-body model images, half-body model images and title images. In some works, shop images resemble high quality street images \cite{Huang2015, Liu2016a}. We classify the state-of-the-art in Table~\ref{tab:state-of-art} following the types of images used in both domains according to three factors: \textit{focus} of the image (full-body model image, half-body model image and title image), \textit{background} (yes/no), and the assumptions made. Note that images from domain 2 are typically professional, while images from domain 1 that contain background can be amateur shots.\\

\paragraph{Approaches.} Early works on street-to-shop or more generally article retrieval \cite{Wang2011b, Fu2012, Liu2012, Kalantidis2013, Yamaguchi2013} are based on classical computer vision: body part detection and/or segmentation, and hand-crafted features. For example, \cite{Liu2012} locate 20 upper and 10 lower body parts and extract HOG \cite{Dalal2005}, LBP \cite{Ojala2002} and colour features for each body part, while \cite{Yamaguchi2013} use clothes parsing to segment their images. In recent years, attention in the fashion-recognition domain has shifted towards using deep learning methods.
Multiple recent studies follow a similar paradigm: the feature representation of both types of images is learnt via attribute classification, while the domain transfer is performed via a ranking loss which pushes matching pairs to have higher scores than non-matching pairs \cite{Huang2015, Wang2016, Liu2016a, Simoserra2016}. In this context, siamese architectures seem quite natural and have given promising results \cite{Huang2015, Wang2016}. When no domain transfer is needed, \ie when query images and assortment images are of the same kind, a single network for the two branches performs just as well \cite{Liu2016a}.

\section{DATA}
\label{sec:data}

\paragraph{Images.} Our dataset contains images of articles from the categories "dress", "jacket", "pullover", "shirt", "skirt", "t-shirt/top" and "trouser" that were sold roughly between 2012 and 2016. The distribution of categories is 35\% t-shirts/tops, 20\% trousers, 14\% pullovers, 14\% dresses, 12\% shirts, 3\% skirts and 2\% jackets. Note that our data is restricted to the categories aforementioned not because of limitations of our approach, but because the selected categories contain the most samples and are likely to be visible. As an example, most of our shoes and socks have no dedicated model images and socks are not visible on other model images.

We have approximately 1.15 million query images in size 224x155 pixels with neutral backgrounds of the type shown in Figures~\ref{fig:images}(a-d). About 250.000 of those are annotated with several articles, while the others are annotated with one article only, even if several articles can be seen. In addition to these query images, our dataset contains approximately 300000 title images in size 224x155 pixels (see Figure~\ref{fig:images}(e)). These 300000 images represent the assortment we want to retrieve from.
\begin{table*}[htb!]
  \begin{center}
    \caption{Variations of the main architecture, the features used, and the matching method in our evaluation.}
	\resizebox{\textwidth}{!}{%
	\begin{tabular}{llllll} \toprule
                  			& global architecture	& features	& query-article-matching	& loss				& inspiration \\ \midrule
          static-fc14-linear		& static		& fc14		& linear			& none  			& \\
          static-fc14-non-linear	& static		& fc14		& non-linear			& cross-entropy			& \cite{Kiapour2015}\\ 
          fc14-non-linear		& ours			& fc14		& non-linear			& cross-entropy			& \\
          \fDNA-linear			& ours			& \fDNA         & linear			& cross-entropy 		& \\
          \fDNA-ranking-loss		& ours			& \fDNA		& linear			& triplet ranking		& \\ 
          all-in-two-nets		& siamese		& learnt	& linear			& triplet ranking		& \cite{Huang2015, Wang2016}\\
					&			&		&        			& + attributes cross-entropy	& \\
          \longname			& ours			& \fDNA		& non-linear			& cross-entropy			& \\
  	  \bottomrule
	\end{tabular}}
    \label{tab:variations}
  \end{center}
\end{table*}

\paragraph{\FashionDNA.} \FashionDNA\ (\fDNA) is \zalando's feature representation for fashion articles. These features are obtained by extracting the activations of a hidden fully connected layer in a static deep convolutional neural network that was trained using title images and article attributes. In our case, these activations are of size 1536, which we reduce to 128 using PCA to decrease the number of parameters in our model. \fDNA\ is not within the scope of this study, but full details about the architecture of the network and the attributes used are given in Appendix.

Because our article features are based on title images, we are slightly different to the other papers which allow model images in their shop images. Our task involves complete domain transfer, while their task is more related to image similarity, especially in the case of \cite{Liu2016a}. This makes comparisons rather difficult. Nevertheless, it was possible to isolate title images in two external datasets which we used as additional test sets.\\

\paragraph{Attributes.} The siamese method implemented in this study requires article attributes. We used category (7 values, available for 100\% of articles), main colour (82 values, 100\% coverage), pattern (19 values, 53\% coverage), clothing length (12 values, 50\% coverage), sleeve length (9 values, 32\% coverage), shirt collar (27 values, 30\% coverage), neckline (12 values, 23\% coverage), material construction (14 values, 20\% coverage), trouser rise (3 values, 14\% coverage).

\section{MATCHING STUDIO PHOTOS TO FASHION ARTICLES OF A GIVEN ASSORTMENT}
\label{sec:method}

\subsection{\longname}
\label{sec:ours}

\paragraph{Building on existing feature representations.} Most recent deep learning approaches to street-to-shop \cite{Huang2015, Wang2016, Liu2016a} can be summarised by Figure~\ref{fig:lower}a. In a forward pass, the left leg of the network processes query images $I_i$ (query-feature module), the right leg processes target images $A_j$ (target-feature module), and a query-article-matching submodel matches the two feature vectors produced. Both inputs are images, and the two legs are trained with varying amounts of weights shared between them (0\% sharing for \cite{Huang2015}, 100\% for \cite{Liu2016a}). On the other end of the spectrum, \cite{Kiapour2015} use static features on both sides, as sketched in Figure~\ref{fig:lower}b.

In contrast, \longname~builds on top of existing feature representations of fashion articles, for example \fDNA. This means that the right leg of the network, as shown in Figure~\ref{fig:lower}c, takes as input the pre-computed features of the articles given their title images.

\paragraph{Non-linear matching of query features to article features.} Most street-to-shop methods relying on an architecture of the type shown in Figure~\ref{fig:lower}a use a triplet ranking loss directly on the feature vectors produced \cite{Huang2015, Wang2016, Liu2016a}.

In contrast, \longname\ follows \cite{Kiapour2015} and uses a more sophisticated submodel. We concatenate the two feature vectors, add on top a batch normalisation layer followed by two fully connected layers with 256 nodes and ReLU activations and one logistic regression layer. We have not come across this combination of a feature learning module and a deep matching module in the literature.

\paragraph{Details of the query-feature submodel.} In the left leg of our network, for the CNN denoted in Figure~\ref{fig:lower}c, we use as a basis the publicly available VGG16 \cite{Simonyan2014} pre-trained on ImageNet ILSVRC-2014 \cite{ImageNet}, but only keep the 13 convolutional layers. On top of these layers we add two fully connected layers with 2048 nodes and ReLU activations, each followed by a 50\% dropout. We finally add a fully connected layer of size 128 which outputs the \fDNA\ of the input query image. All details are given in Appendix.

\paragraph{Backward pass.} Approaches of the type shown in Figure~\ref{fig:lower}a \cite{Huang2015, Wang2016, Liu2016a} use article meta-data or attributes on each side of the two-legged network. $\theta$ denotes the joint set of parameters of the two legs. On top of the layer generating the feature vector $f(.|\theta)$ sits, for each attribute, a sub-network ending with softmax activations. A categorical cross-entropy loss is used for each attribute-specific branch. Finally a triplet ranking loss joins the two feature representations.

In contrast, we disregard attributes and use a cross-entropy loss for our query-article-matching submodel. The loss is given in Equation~\ref{eq:loss}, where $N$ is the number of query images, $M$ the number of articles, $p_{ij}$ the output of the model given the input query image $I_i$ and target article $A_j$, and $y_{ij}$ is the ground truth label: 1 if the target article $A_j$ belongs to the query model image $I_i$, 0 otherwise. Note that the loss is back-propagated to all layers of the network, even to layers that were pretrained on ImageNet.
\begin{equation}
\mathcal{L} = \displaystyle{\sum_{i=1}^{N}{\sum_{j=1}^{M}{y_{ij} \log{p_{ij}} + \left(1 - y_{ij}\right) \log{\left(1 - p_{ij}\right)}}}}
\label{eq:loss}
\end{equation}
For practical reasons, we split our query images in mini-batches of size 64, and compare them with 50 articles only. These 50 articles contain the (at least 1) annotated matching articles and (at most 49) articles drawn randomly without replacement and temporarily labelled as negative matches. The negative articles are constantly resampled (for each query image in the batch, for each batch, and for each epoch), which enforces diversity. Our images are not fully annotated and some of the data supplied might be erroneously labelled as negative, however with so many articles available and so few articles actually present in query images, the probability that such an incorrect negative label occurs is relatively low. The loss over a mini-batch is shown in Equation~\ref{eq:batch-loss}, where $c(i)$ is the $i^{\text{th}}$ query image in the batch, and $c(i,j)$ the $j^{\text{th}}$ article for this image in the batch.
\begin{equation}
  \begin{split}
    \mathcal{L} = \sum_{i=1}^{64} \sum_{j=1}^{50} & y_{c(i)c(i,j)} \log{p_{c(i)c(i,j)}} \\
                                                  & + \left(1 - y_{c(i)c(i,j)}\right) \log{\left(1 - p_{c(i)c(i,j)}\right)}
  \end{split}
  \label{eq:batch-loss}
\end{equation}

\subsection{Other Approaches}
\label{sec:competitors}

We compare ourselves with several alternatives, some of which are inspired by the literature.
\begin{itemize}
  \item As an alternative to \fDNA, we also use features extracted from layer fc14 of a VGG16 network pre-trained on ImageNet with their dimensionality reduced to 128 by PCA for comparability (denoted fc14), and from \cite{Simoserra2016} (denoted 128floats).
  \item We implement the architecture variations mentioned in Figure~\ref{fig:lower}.
  \item We implement several alternative query-article-matching and loss strategies. The various combinations are listed in Table~\ref{tab:variations}.
\end{itemize}

In this study, the triplet ranking loss is given by $l_{ijk} = \sigma\left(f(I_i|\theta)^T\ \left(f(A_k|\gamma) - f(A_j|\gamma)\right)\right)$ where $I_i$ is a query image, $A_j$ an article present in $I_i$, $A_k$ an article which is not known to be present in $I_i$, $\theta$ the set of parameters of the model, $\gamma$ is $\theta$ for a siamese model and empty otherwise, $f(.|.)$ the feature vector of an image given parameters and $\sigma$ the sigmoid function. For each query image of each batch, one positive article is sampled as $A_j$, and 50 negative articles are sampled as $A_k$, giving 50 triplets at a time. The 50 losses are summed, as shown in Equation \ref{eq:batch-tloss}, where $c(i)$ is the $i^{\text{th}}$ query image in the batch, $c(i,+)$ the sampled positive article for this image, and $c(i,j)$ the $j^{\text{th}}$ negative article.\footnotemark[1]
\begin{equation}
  \mathcal{L} = \sum_{i=1}^{64} \sum_{j=1}^{50} l_{c(i)c(i,+)c(i,j)}
  \label{eq:batch-tloss}
\end{equation}

In our all-in-two-nets model, the right leg has the same architecture as the left leg described in Section~\ref{sec:ours} and is also pretrained on ImageNet, but no parameters are shared between the two legs, as in \cite{Huang2015}. All layers are trainable, even the pretrained ones.

\section{RESULTS}
\label{sec:evaluation}

\subsection{Retrieval on Test Query Images}

\paragraph{Experimental set-up.} We randomly split the dataset into training and test set, with 80\% of query images and articles being kept for training. Many of our full-body images are annotated with several articles that may be shared across images, making a clean split of articles impossible. To avoid using training articles at test time, we discard them from the retrieval set, and discard query images that were annotated with such articles. As a result, our pool of test queries is slightly biased towards half-body images.

We run tests on 20000 randomly sampled test query images against 50000 test articles, which is roughly the number of articles of the selected categories at a given time in \zalando's assortment. For each test query image, all 50000 possible (image, article) pairs are submitted to the model and are ranked according to their score.
\begin{table*}[htb!]
  \begin{center}
  \caption{Results of the retrieval test using 20000 query images against 50000 \zalando\ articles. Top-$k$ indicates the proportion of query images for which the correct article was found at position $k$ or below. Average and median refer respectively to the average and median position at which an article is retrieved. The best performance is shown in bold. A plot of these results can also be found in Figure~\ref{fig:topk} in Appendix.}
  \resizebox{\textwidth}{!}{%
  \begin{tabular}{lccccccrr} \toprule
    				& top-1 & top-5 & top-10 & top-20 & top-50 & top-1\% & average & median \\ \midrule 
      static-fc14-linear	& 0.005 & 0.015 & 0.022  & 0.033  & 0.054  & 0.166   & 13502   & 8139   \\
      static-fc14-non-linear	& 0.030 & 0.083 & 0.121  & 0.176  & 0.271  & 0.609   & 1672    & 258    \\
      fc14-non-linear		& 0.131 & 0.317 & 0.423  & 0.539  & 0.684  & 0.926   & 230     & 15     \\
      128floats-non-linear	& 0.132 & 0.319 & 0.426  & 0.540  & 0.677  & 0.909   & 274     & 15     \\
      \fDNA-ranking-loss	& 0.091 & 0.263 & 0.372  & 0.494  & 0.660  & 0.933   & 177     & 20     \\
      \fDNA-linear		& 0.121 & 0.314 & 0.423  & 0.547  & 0.700  & 0.936   & 178     & 15     \\
      all-in-two-nets		& 0.033 & 0.115 & 0.181  & 0.277  & 0.438  & 0.838   & 620     & 69    \\
      \longname		& \textbf{0.238} & \textbf{0.496} & \textbf{0.613} & \textbf{0.722} & \textbf{0.834} & \textbf{0.970} & \textbf{89} & \textbf{5} \\
\bottomrule
  \end{tabular}}
  \label{tab:results}
  \end{center}
\end{table*}
\begin{figure*}[htb!]
  \begin{center}
    \includegraphics[width=\textwidth]{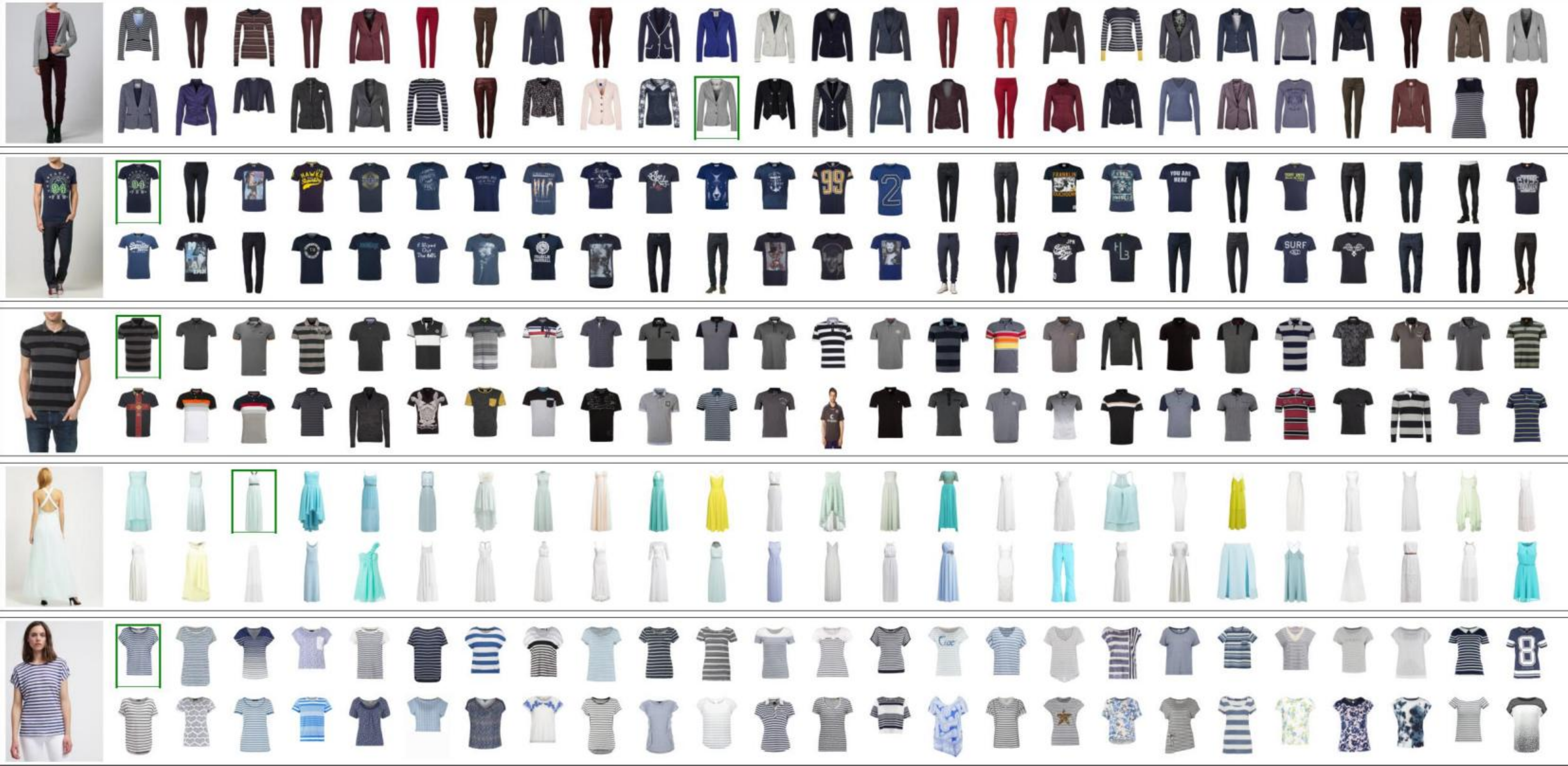}
    \caption{Random examples of the retrieval test using 20000 queries against 50000 \zalando\ articles. Query images are in the left-most column. Each query image is next to two rows displaying the top 50 retrieved articles, from left to right, top to bottom. Green boxes show exact hits.}
    \label{fig:collages}
  \end{center}
\end{figure*}

\paragraph{Retrieval performance.} Table~\ref{tab:results} shows the performance of the various models (a plot of these results can also be found in Figure~\ref{fig:topk} in Appendix). We use as performance measure top-$k$ retrieval, which gives the proportion of query images for which the correct article was found at position $k$ or below. We add a top-1\% measure which here means top-500, so that research groups with a different number of articles may compare their results to ours more easily. Average refers to the average position (or rank) at which an article is retrieved. We expect the distribution of retrieval positions to be heavy-tailed, so we include the median position as a more robust measure. All our models are assessed on the exact same (query, article) pairs.
\begin{table*}[htb!]
  \begin{center}
  \caption{Results of the retrieval test on DeepFashion In-Shop-Retrieval \cite{Liu2016a} using 2922 query images against 683 articles, and on LookBook \cite{Yoo2016} using 68820 query images against 8726 articles. Top-$k$ indicates the proportion of query images for which the correct article was found at position $k$ or below. Average and median refer respectively to the average and median position at which an article is retrieved.}
  \resizebox{\textwidth}{!}{%
  \begin{tabular}{llcccccccrr} \toprule
    				&				& top-1 & top-5 & top-10 & top-20 & top-50 & top-1\% & average & median \\ \midrule
      DeepFashion 		
      				& fc14-non-linear		& 0.159 & 0.439 & 0.565 & 0.689 & 0.838 & 0.475 & 32   & 6 \\ 
				& \longname			& 0.258	& 0.578 & 0.712 & 0.818 & 0.919 & 0.619 & 17   & 3 \\ \bottomrule 
      LookBook			
      				& fc14-non-linear		& 0.009 & 0.031 & 0.050 & 0.078 & 0.134 & 0.181 & 1657 & 797 \\ 
      				& \longname			& 0.013 & 0.044 & 0.070 & 0.107 & 0.182 & 0.241 & 1266 & 466 \\ 
  \bottomrule \end{tabular}}
  \label{tab:external}
  \end{center}
\end{table*}

\longname\ outperforms the other models, with a median article position of 5, mostly because of the non-linear matching module. In general, fc14 with our architecture performs surprisingly well with a median index of 15 and would already be suited for practical applications, which supports our one-leg approach. It is surpassed however by \fDNA, indicating that having a fashion-specific representation does help. 128floats is pretrained on ImageNet and finetuned using a fashion dataset, but this dataset seems too limited in size to make a difference with fc14 on this task.

A second observation is that learning our feature representation from scratch does not perform as well as a pre-trained feature representation. The reason is two-fold. Firstly, \fDNA\ was trained on many more articles than we have in this study, including articles of other categories such as shoes for example, or swimming suits. Secondly, having a pre-trained feature representation heavily reduces the model complexity, which is desirable in terms of computational time, but also in the presence of limited datasets. Extensive architecture exploration may lead to a siamese architecture that outperforms \longname. We do not conclude that siamese architectures are not performant, only that a one-leg architecture is a viable alternative.

A third observation is that a ranking loss does not seem necessary. In our case, it performed worse than the cross-entropy loss. It is also slower to train as it requires triplets of images instead of pairs and, because it is trained to rank and not to predict a match, the scores produced are meaningless.

The total time needed for (naive) retrieval using \longname\ is given in Figure~\ref{fig:timings} in Appendix. While features generated by the query-feature submodel can be pre-computed for efficiency, the query-article-matching submodel still needs to be run and its non-linearity could be a limitation in a realistic scenario. A possibility could be to use \fDNA-linear (which achieves the second-best performance and is extremely fast since only a dot product is needed) to identify a subset of candidate articles on which it is worth running the non-linear match of \longname.

\paragraph{Article retrieval.} Figure~\ref{fig:collages} shows randomly selected examples of retrievals, with the query image shown on the left, followed by the top 50 articles. Even if the article is not retrieved at the top position, style is conserved, suggesting that the model has indeed generalised beyond simple attributes. Moreover, we are able to find more than one category when they are sufficiently visible, for example similar trousers are also found early in the upper two images, and in the top image t-shirts also appear, though there seems to be some degree of confusion as to which colour belongs to which garment.

\subsection{Results on External Datasets}

\paragraph{The datasets.} To assess the usability of our model, we run tests on two external datasets, namely \emph{DeepFashion In-Shop-Retrieval} \cite{Liu2016a}, which is the closest to ours and contains 683 title images and 2922 matching shop images, and \emph{LookBook} \cite{Yoo2016}, which contains 8726 title images and 68820 matching shop images with backgrounds. For both datasets, the background is different to ours, and the aspect ratio of the people in query images may also vary due to image resizing. As a result, the image distributions deviate from ours, both for queries and for targets. Note that publicly available datasets such as Exact-Street-To-Shop \cite{Kiapour2015} or DeepFashion  \cite{Liu2016a} have a mixture of image types as targets and can therefore not be used directly. Nevertheless, for DeepFashion In-Shop-Retrieval and Lookbook, it was easily possible to isolate title images and to restructure the dataset for our needs.

LookBook was originally not meant for street-to-shop, many articles have more than one ID and are treated as different so we are not rewarded for finding them with the wrong ID. The images are very different from ours and we do not really expect any good results, but we use it to test the boundaries of DeepArticleFinder.
\begin{figure*}[htb!]
  \begin{center}
    \includegraphics[width=\textwidth]{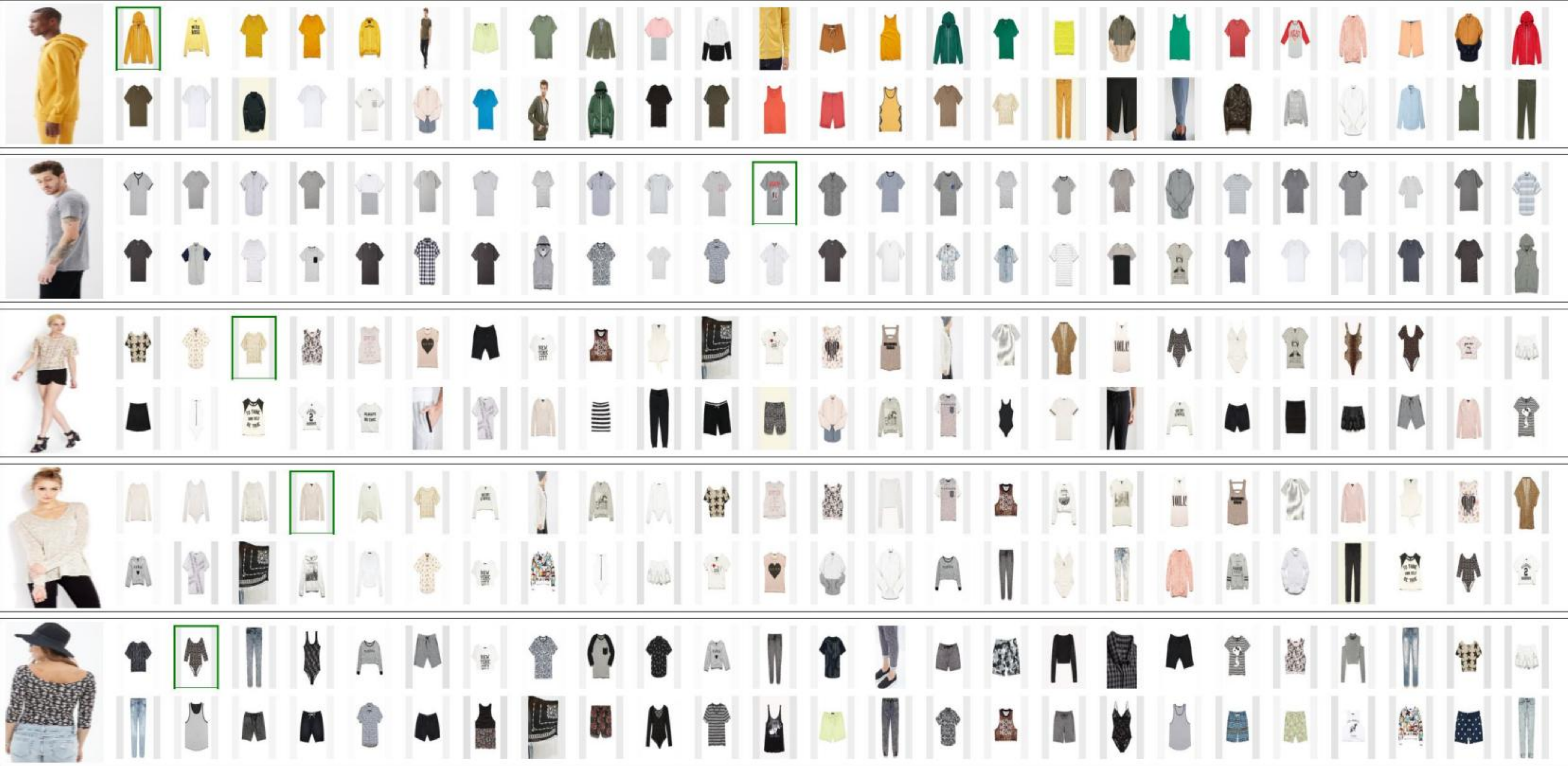}
    \caption{Random examples of outcomes of the retrieval test on query images from DeepFashion In-Shop-Retrieval \cite{Liu2016a}. Query images are in the left-most column. Each query image is next to two rows displaying the top 50 retrieved articles, from left to right, top to bottom. Green boxes show exact hits.}
    \label{fig:deepfashion_quantitative_collages}
  \end{center}
\end{figure*}
\begin{figure*}[htb!]
  \begin{center}
    \includegraphics[width=\textwidth]{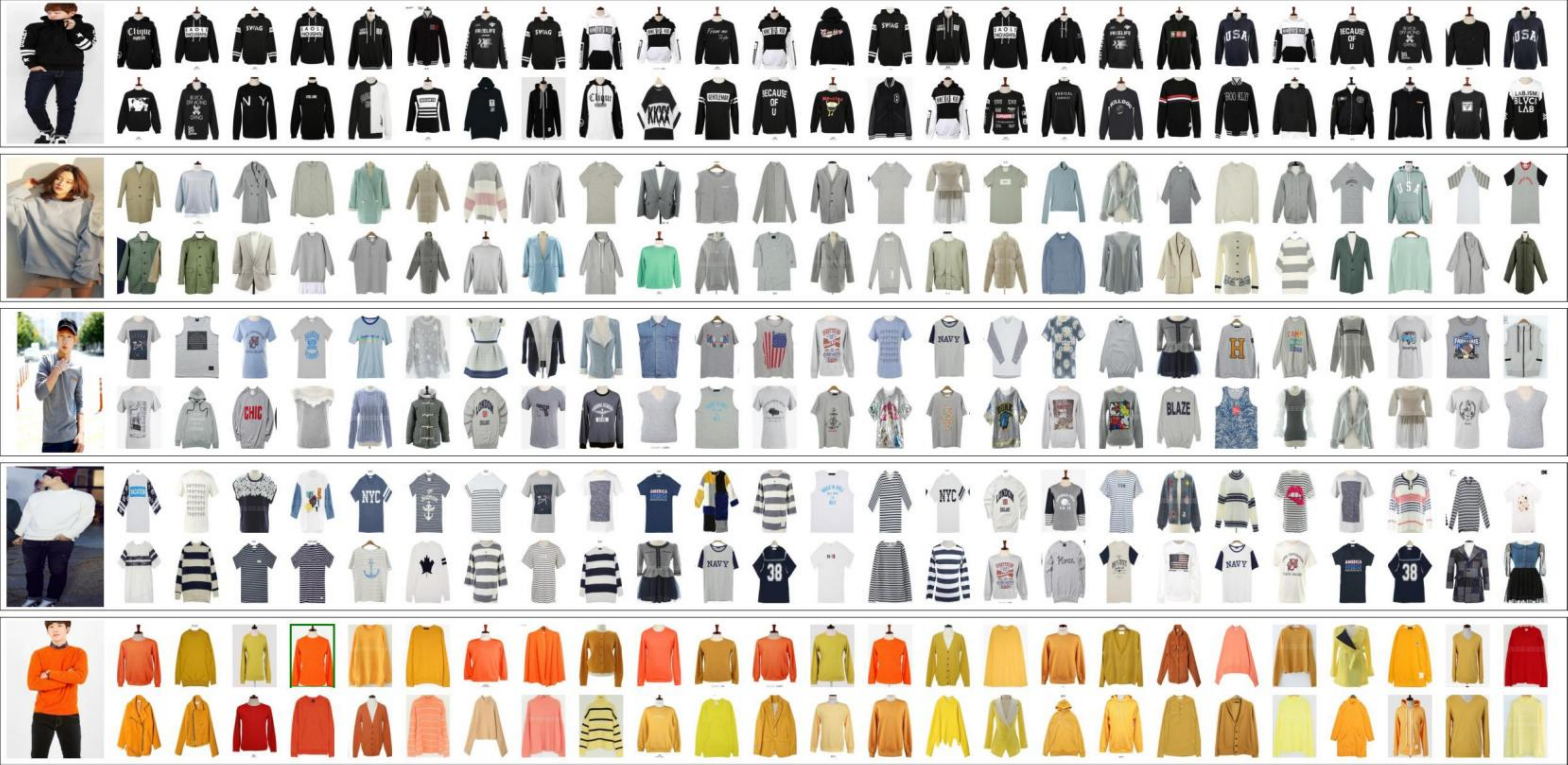}
    \caption{Random examples of outcomes of the retrieval test on query images from LookBook \cite{Yoo2016}. Query images are in the left-most column. Each query image is next to two rows displaying the top 50 retrieved articles, from left to right, top to bottom. Green boxes show exact hits.}
    \label{fig:lookbook_quantitative_collages}
  \end{center}
\end{figure*}
\begin{figure*}[htb!]
  \begin{center}
    \includegraphics[width=\textwidth]{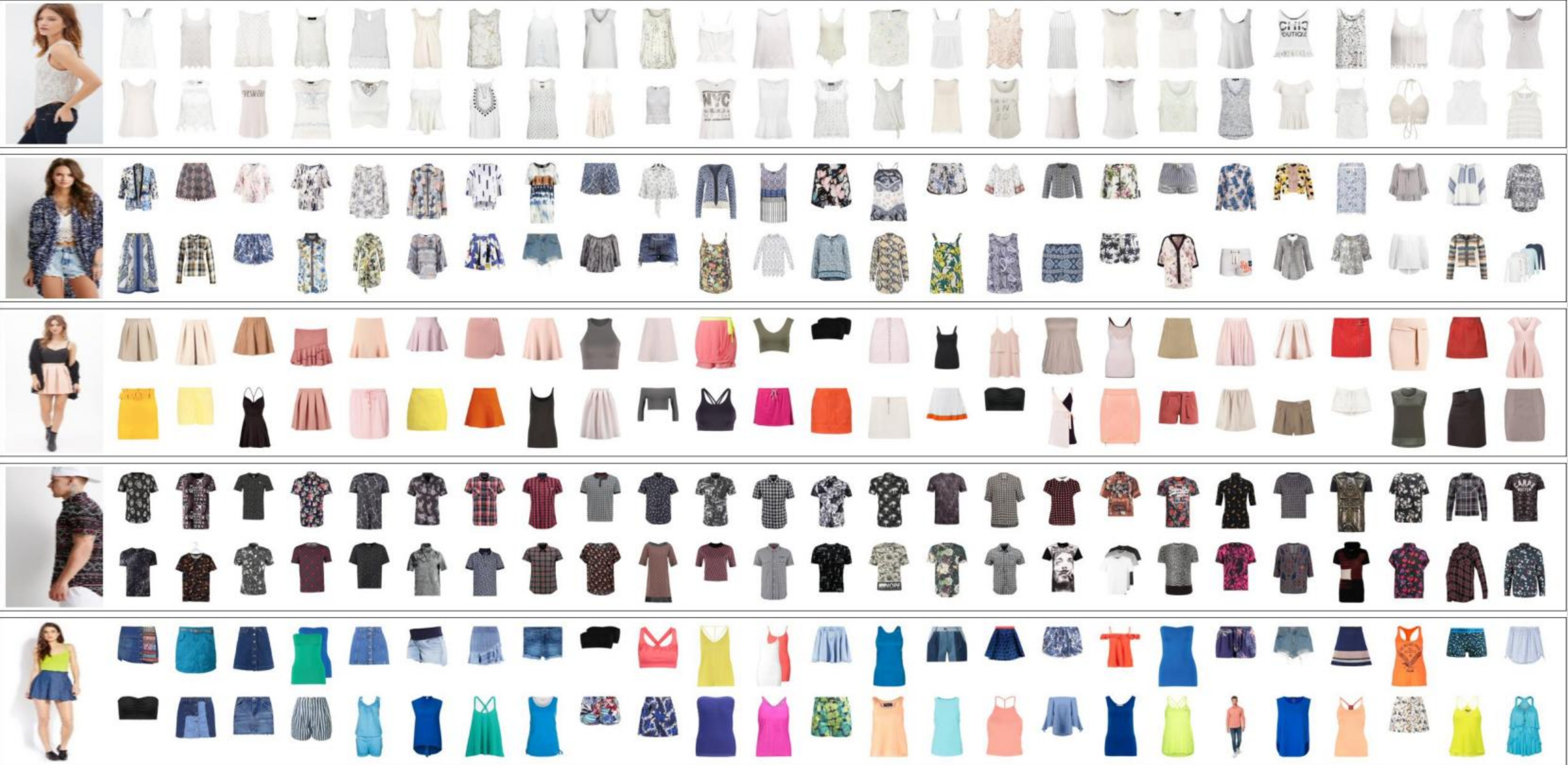}
    \caption{Random examples of outcomes of the retrieval test on query images from DeepFashion In-Shop-Retrieval \cite{Liu2016a} against 50000 \zalando\ articles. Query images are in the left-most columns. Each query image is next to two rows showing the top 50 retrieved articles, from left to right, top to bottom.}
    \label{fig:deepfashion_qualitative_collages}
  \end{center}
\end{figure*}
\begin{figure*}[htb!]
  \begin{center}
    \includegraphics[width=\textwidth]{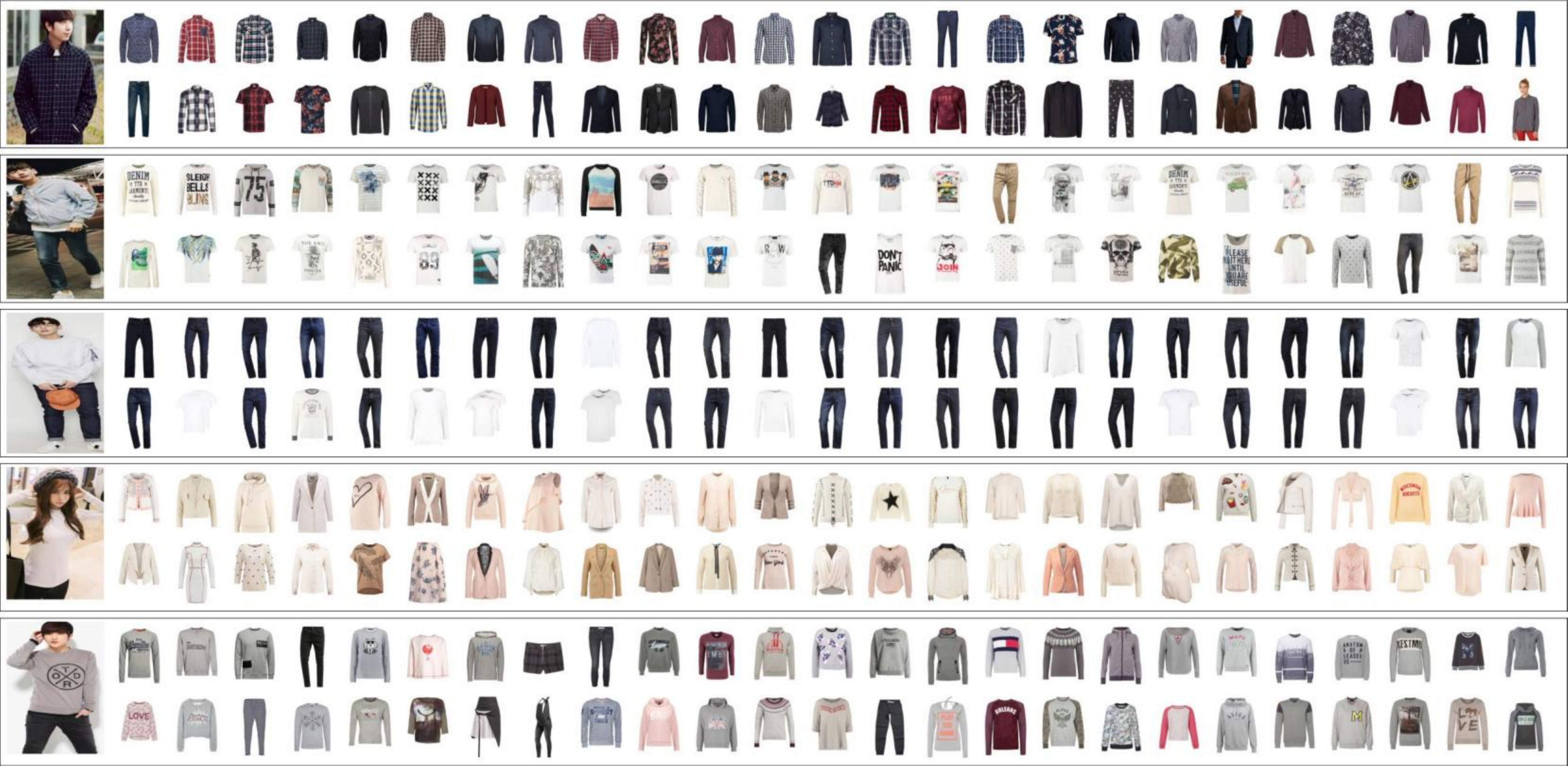}
    \caption{Random examples of outcomes of the retrieval test on query images from LookBook \cite{Yoo2016} against 50000 \zalando\ articles. Query images are in the left-most columns. Each query image is next to two rows showing the top 50 retrieved articles, from left to right, top to bottom.}
    \label{fig:lookbook_qualitative_collages}
  \end{center}
\end{figure*}

\paragraph{Retrieval test.} We compute the static features of the title images using the pre-existing feature representation of interest (fc14 or \fDNA) and simply apply our model to their query images. It is critical to note here that our model is not fine-tuned to this new data. The results are shown in Table~\ref{tab:external}. Figures~\ref{fig:deepfashion_quantitative_collages} and~\ref{fig:lookbook_quantitative_collages} show the retrieval outcomes for 5 randomly selected query images from DeepFashion In-Shop-Retrieval and LookBook respectively. Note that, in DeepFashion In-Shop-Retrieval, where full bodies can be seen, more than one category can be retrieved. LookBook has the added difficulty of containing backgrounds. However, when these remain understated, our model can find adequate suggestions.

\paragraph{Finding similar articles in \zalando's assortment.} While the previous tests allow us to assess performance, in practice we would like to suggest articles from our own assortment. We run qualitative tests on query images from DeepFashion In-Shop-Retrieval and LookBook using our own test articles as retrieval set. Figures~\ref{fig:deepfashion_qualitative_collages} and~\ref{fig:lookbook_qualitative_collages} show the retrieval outcomes for 5 randomly selected query images from DeepFashion In-Shop-Retrieval and LookBook respectively, and indicate that our model can make very appropriate suggestions for DeepFashion In-Shop-Retrieval, and to a certain extent for LookBook when background is understated.

\section{CONCLUSION}

We have presented \longname, a model for article recognition in fashion images with neutral backgrounds. Instead of solving the problem from scratch as most recent studies have done, \longname\ builds on top of existing feature representations for fashion articles, 
and projects query images onto this fixed feature space using a deep convolutional neural network. We show that our approach is most often able to find correct articles within a range that is sufficiently small for building realistic visual search engines in the studio-to-shop setting.

Our method is easy to implement and only requires article features and matches between query images and articles. We find that we achieve satisfactory results without having to use additional meta-data for our query images or articles, and that, in the absence of specific features such as \FashionDNA, the publicly available features fc14 or 128floats are already quite powerful.

While our dataset does not contain street images, many of the obstacles in computer vision for fashion such as occlusion and deformation remain, yet our results are very promising. We are currently working on extending our approach to street images, making use of state-of-the-art image segmentation techniques and external datasets.

\section*{\uppercase{Acknowledgements}}

\noindent The authors would like to thank \seb\ and \christian\ for their help with \FashionDNA.

\bibliographystyle{apalike}
{\small
\bibliography{ms}}

\section*{\uppercase{Appendix}}

\subsection*{\FashionDNA: Implementation Details}

\fDNA\ features are extracted from the deep convolutional neural network described in Table~\ref{tab:fDNA}. This net is trained on title images to predict a number of article attributes, summarised in Table~\ref{tab:fdna_attributes}. Training is done via multiple categorical cross-entropy losses, one for each attribute.
\begin{table}[htb!]
  \begin{center}
  \caption{Architecture of the network used for the \fDNA\ features of title images.}
  \resizebox{\columnwidth}{!}{%
    \begin{tabular}{llrrrr}\toprule
      name		& type		& kernel	& activation	& output size	& \# params \\\midrule
			& InputLayer	&		&		& 256x177x3	& 0 \\
			& Conv2D	& 11x11		& ReLU		& 62x42x96	& 34.944 \\
			& MaxPooling2D  &		&		& 31x21x96	& 0 \\
			& LRN		&		&		& 31x21x96	& 0 \\
			& Conv2D	& 5x5		& ReLU		& 31x21x256	& 614.656 \\
			& MaxPooling2D	&		&		& 15x10x256	& 0 \\
			& LRN		&		&		& 15x10x256	& 0 \\
			& Conv2D	& 3x3		& ReLU		& 15x10x384	& 885.120 \\
			& Conv2D	& 3x3		& ReLU		& 15x10x384	& 1.327.488 \\
			& Conv2D	& 3x3		& ReLU		& 15x10x256	& 884.992 \\
			& MaxPooling2D	&		&		& 7x5x256	& 0 \\
			& Flatten	&		&		& 8960		& 0 \\
      \fDNA		& Dense		&		& ReLU		& 1536		& 13.764.096 \\ \midrule
			& Dense		&		& ReLU		& 1024		& 1.573.888 \\
      commodity		& Dense		&		& softmax	& 1448		& 1.484.200 \\ \midrule
			& Dense		&		& ReLU		& 890		& 1.367.930\\
      article\_number	& Dense		&		& softmax	& 445		& 396.495 \\ \midrule
			& Dense		&		& ReLU		& 160		& 245.920 \\
      silhouette	& Dense		&		& softmax	& 80		& 12.880 \\ \midrule
			& Dense		&		& ReLU		& 1024		& 1.573.888 \\
      brand		& Dense		&		& softmax	& 3719		& 3.811.975 \\ \midrule
			& Dense		&		& ReLU		& 306		& 470.322 \\
      target\_group	& Dense		&		& softmax	& 153		& 46.971 \\  \midrule
			& Dense		&		& ReLU		& 40		& 61.480 \\
      pattern		& Dense		&		& softmax	& 19		& 779 \\  \midrule
			& Dense		&		& ReLU		& 72		& 110.664 \\ 
      material		& Dense		&		& softmax	& 36		& 2.628\\ \midrule
			& Dense		&		& ReLU		& 244		& 375.028 \\
      main\_colour	& Dense		&		& softmax	& 122		& 29.890 \\  \midrule
			& Dense		&		& ReLU		& 140		& 215.180 \\
      second\_colour	& Dense		&		& softmax	& 70		& 9870 \\
      \bottomrule	&		&		&		&		& 29.301.284 \\
    \bottomrule\end{tabular}}
  \label{tab:fDNA}
  \end{center}
\end{table}

\begin{table}[htb!]
  \centering
  \caption{Attributes used for training \FashionDNA. Commodity group, statistical article number and silhouette all describe article categories on different levels of granularity. The target group attribute describes combinations of age and gender.}
  \begin{tabular}{lrr} \toprule
    attribute			& \# values	& coverage [\%] \\ \midrule 
    commodity group		& 1448		& 100 \\ 
    statistical article number	& 445		& 76 \\ 
    silhouette			& 80		& 100 \\
    brand			& 3719		& 98 \\
    target group		& 153		& 100 \\
    pattern			& 19		& 1 \\
    material			& 36		& 1 \\
    main colour			& 122		& 100 \\
    second colour		& 70		& 7 \\
  \bottomrule \end{tabular}
  \label{tab:fdna_attributes}
\end{table}

\clearpage
\subsection*{Architecture of \longname}

\begin{table}[htb!]
  \begin{center}
  \caption{Architecture of \longname.}
  \resizebox{0.5\textwidth}{!}{%
    \begin{tabular}{llrrr}\toprule
      name		& type		& activation	& output size	& \# params \\\midrule
      \multicolumn{5}{l}{\textit{query-feature submodule}} \\ 
      query\_input	& InputLayer	&		& 224x155x3	& 0 \\
	                & VGG16		&		& 7x4x512	& 14,714,688 \\
			& Flatten	&		& 14436		& 0 \\
			& Dense		& ReLU		& 2048		& 29,362,176 \\
			& Dropout(0.5)	&		& 2048		& 0 \\
			& Dense		& ReLU		& 2048		& 4,196,352 \\
			& Dropout(0.5)	&		& 2048		& 0 \\
   query\_\fDNA	& Dense		&		& 128		& 262,272 \\ \midrule
      \multicolumn{5}{l}{\textit{query-article-matching submodule}} \\ 
     article\_\fDNA	& InputLayer	&		& 128		& 0 \\ 
    query\_\fDNA	& InputLayer	&		& 128		& 0 \\ 
     			& Concatenate	&		& 256		& 0 \\
			& BatchNorm	&		& 256		& 512 \\
			& Dense		& ReLU		& 256		& 65792 \\
			& Dense		& ReLU		& 256		& 65792 \\
			& Dense		& sigmoid	& 1		& 257 \\
      \bottomrule	&		&		&		& 48,667,841 \\
    \bottomrule\end{tabular}}
  \label{tab:net}
  \end{center}
\end{table}

\subsection*{Top-$k$ Retrieval Results}

\begin{figure}[htb!]
  \begin{center}
    \includegraphics[width=\linewidth]{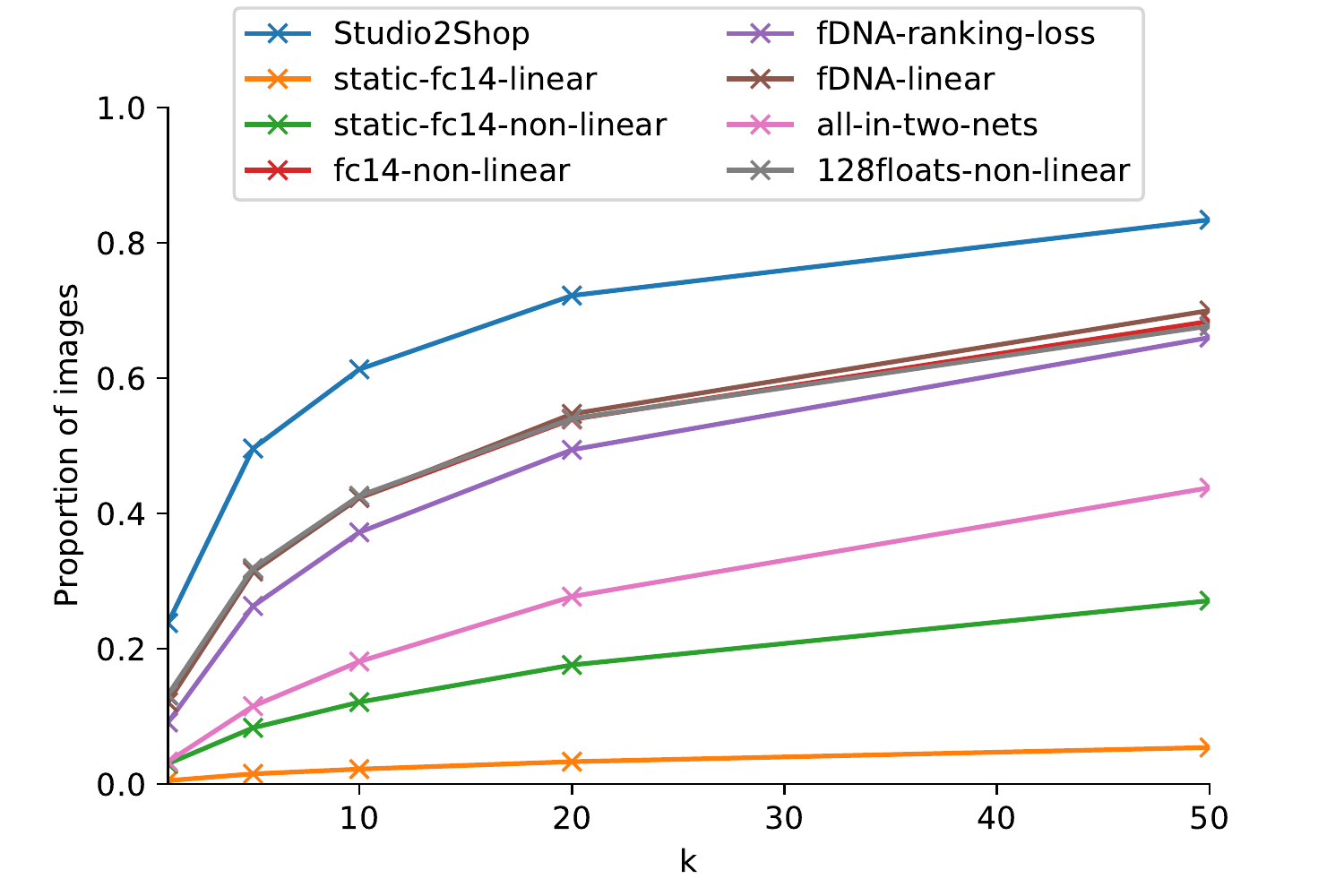}
    \caption{Top-$k$ results of the retrieval test using 20000 query images against 50000 \zalando\ articles. The top-$k$ metric gives the proportion of query images for which the correct article was found at position $k$ or below.}
    \label{fig:topk}
  \end{center}
\end{figure}

\subsection*{Timings}

\begin{figure}[htb!]
  \begin{center}
    \includegraphics[width=\linewidth]{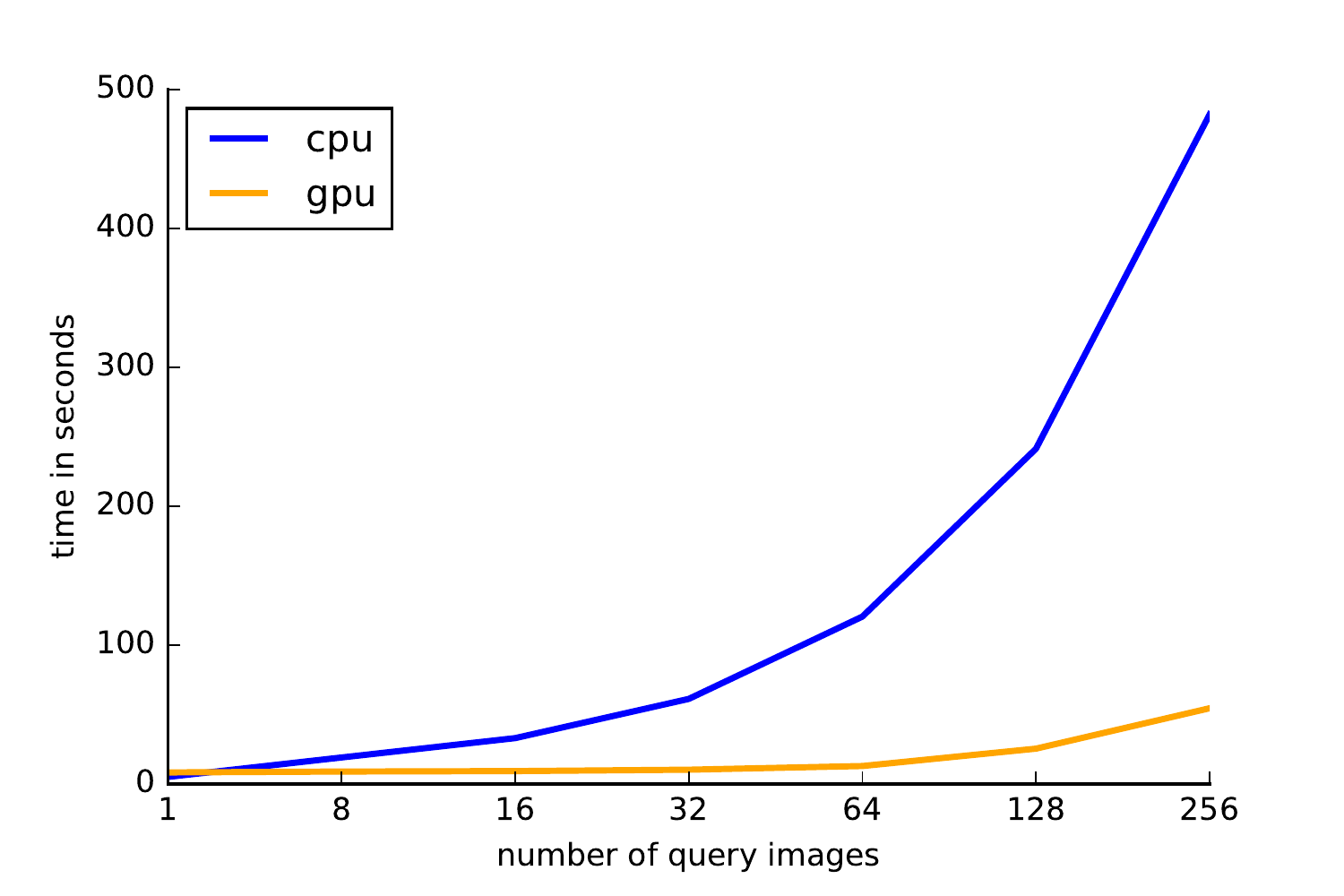}
    \caption{Time needed to test query images against 50000 articles with a fixed batch size of 64 and using 50 articles at a time. Model loading is not part of the time calculation. The experiments are run from scratch (apart from model loading) for each number of images and averaged over 10 repetitions. To process only one image, it takes about 5s on a CPU and 8s on a GPU.}
    \label{fig:timings}
  \end{center}
\end{figure}

Figure~\ref{fig:timings} shows the time needed, both for a CPU (Intel Xeon processor with 3.5 GHz) and for a GPU (Nvidia K80), to test query images against 50000 articles with a fixed batch size of 64, which is rather small but common. First, the features of a query image are extracted using the query-feature submodel, then the query-article-matching submodel is applied to all images against all the articles, using 50 articles at a time. These two steps (query-feature + query-article-matching) are part of the calculation, however model loading and image resizing are not. These numbers are rough but are averaged over 10 repetitions and are meant to give an order of magnitude.

To process only one image, it takes about five seconds on a CPU, 8 on a GPU (the difference is explained by the overhead inherent to GPU calculations that becomes very quickly negligible as the number of images increases). Note that our implementation is not optimised for speed. For example, there is no parallelisation. Moreover, we use the exact same architecture as for training and therefore only test against 50 articles at a time. For production, retrieval time could be substantially decreased by testing instead against hundreds of articles at a time. Further speed-ups could be achieved by increasing the batch size when running on a GPU.

\newpage
\vfill
\end{document}